\documentclass[journal,twoside,web]{ieeecolor}
\usepackage{tmi}
\usepackage{cite}
\usepackage{amsmath,amssymb,amsfonts}
\usepackage{algorithmic}
\usepackage{graphicx}
\usepackage{textcomp}
\usepackage{array}
\usepackage{booktabs}
\usepackage{multirow}
\usepackage{xcolor}

\usepackage{soul}

\def\BibTeX{{\rm B\kern-.05em{\sc i\kern-.025em b}\kern-.08em
    T\kern-.1667em\lower.7ex\hbox{E}\kern-.125emX}}
\begin{document}
\title{MedSeqFT: Sequential Fine-tuning Foundation Models for 3D Medical Image Segmentation}
\author{
Yiwen Ye,
Yicheng Wu*,
Xiangde Luo,
He Zhang,
Ziyang Chen,
Ting Dang,
Yanning Zhang, ~\IEEEmembership{Fellow,~IEEE}
Yong Xia*,~\IEEEmembership{Member,~IEEE}
\thanks{This work was supported by the National Natural Science Foundation of China under Grant 62171377 and Grant 92470101. ({\em Corresponding authors: Y. Wu and Y. Xia.})}
\thanks{Y. Ye, Z. Chen, Y. Zhang, and Y. Xia are with the School of Computer Science and Engineering, Northwestern Polytechnical University, Xi’an, Shaanxi, 710072, China (e-mail: \{ywye, zychen\}@mail.nwpu.edu.cn, \{ynzhang, yxia\}@nwpu.edu.cn).}
\thanks{Y. Wu is with Monash University, Clayton, VIC 3168, Australia (e-mail: yicheng.wu@monash.edu).}
\thanks{X. Luo is with the Department of Radiation Oncology, Sichuan Cancer Hospital, Chengdu, Sichuan, 610041, China (e-mail: luoxd1996@gmail.com).}
\thanks{H. Zhang is with RMIT, Melbourne, VIC 3001, Australia (e-mail: he.zhang@rmit.edu.au).}
\thanks{T. Dang is with the University of Melbourne, Melbourne, VIC 3053, Australia (e-mail: ting.dang@unimelb.edu.au).}
}
\maketitle

\begin{abstract}
Foundation models have become a promising paradigm for advancing medical image analysis, particularly for segmentation tasks where downstream applications often emerge sequentially. Existing fine-tuning strategies, however, remain limited: parallel fine-tuning isolates tasks and fails to exploit shared knowledge, while multi-task fine-tuning requires simultaneous access to all datasets and struggles with incremental task integration. To address these challenges, we propose MedSeqFT, a sequential fine-tuning framework that progressively adapts pre-trained models to new tasks while refining their representational capacity. MedSeqFT introduces two core components: (1) Maximum Data Similarity (MDS) selection, which identifies downstream samples most representative of the original pre-training distribution to preserve general knowledge, and (2) Knowledge and Generalization Retention Fine-Tuning (K\&G RFT), a LoRA-based knowledge distillation scheme that balances task-specific adaptation with the retention of pre-trained knowledge. Extensive experiments on two multi-task datasets covering ten 3D segmentation tasks demonstrate that MedSeqFT consistently outperforms state-of-the-art fine-tuning strategies, yielding substantial performance gains (e.g., an average Dice improvement of 3.0\%). Furthermore, evaluations on two unseen tasks (COVID-19-20 and Kidney) verify that MedSeqFT enhances transferability, particularly for tumor segmentation. Visual analyses of loss landscapes and parameter variations further highlight the robustness of MedSeqFT. These results establish sequential fine-tuning as an effective, knowledge-retentive paradigm for adapting foundation models to evolving clinical tasks.
Code will be released.
\end{abstract}


\begin{figure}[t]
  \centering
  \includegraphics[width=0.95\linewidth]{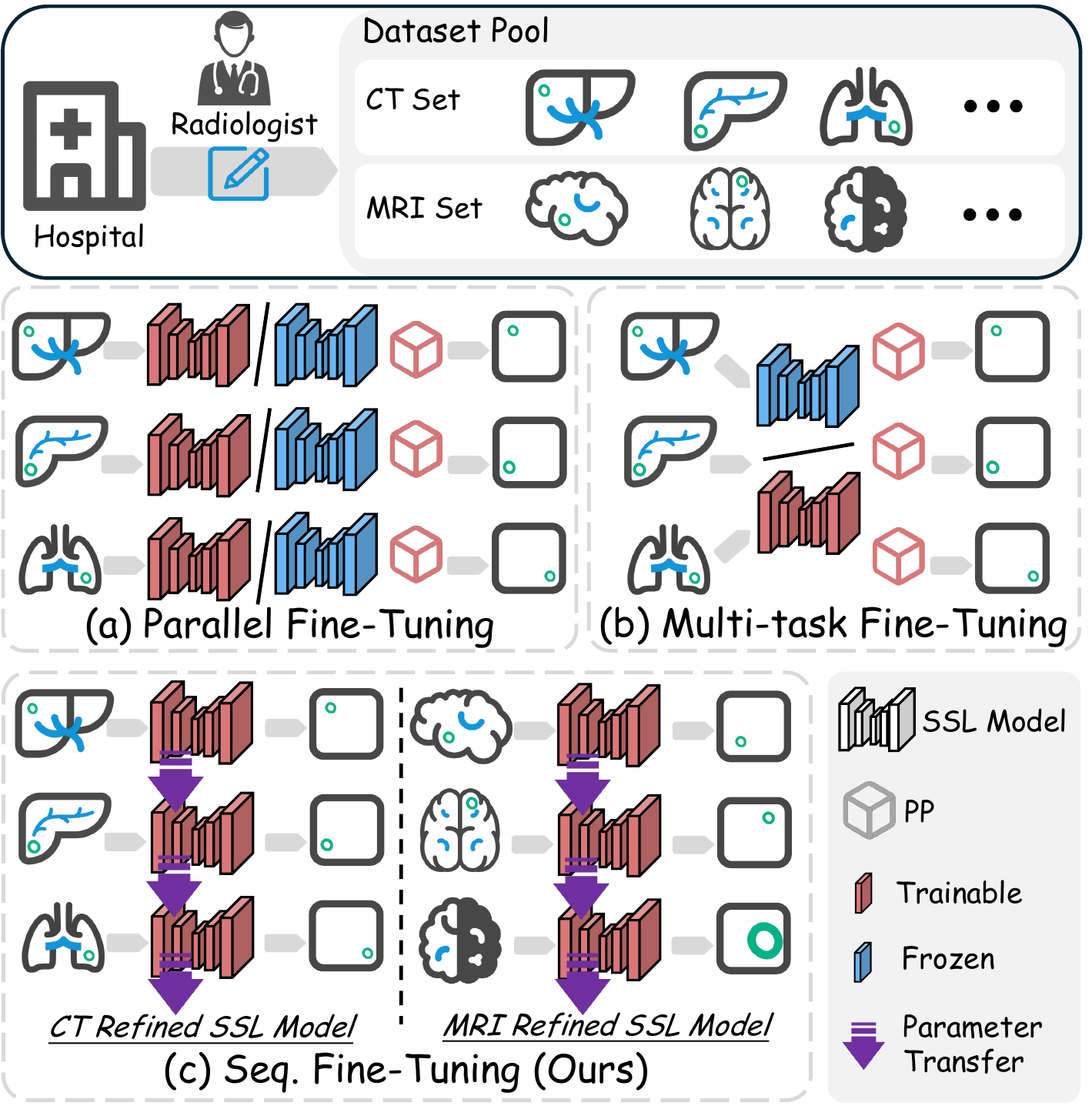}
   \caption{Three strategies for fine-tuning self-supervised learning (SSL) models on multi-task datasets. (a) Parallel fine-tuning: The SSL model $M_0$ is independently fine-tuned on each dataset from the dataset pool, with all or partial parameters (PP) updated. (b) Multi-task Fine-tuning: Multiple tasks are fine-tuned jointly, with $M_0$ shared across tasks, while the PP can either be task-shared or task-independent. (c) Sequential Fine-tuning: $M_0$ is fine-tuned sequentially across datasets. After fine-tuning on the first dataset, it becomes $M_1$, which then serves as the pre-trained model for the next dataset, and so on.
   For all strategies, segmentation heads, which remain learnable, are omitted for clarity.
   }
\label{fig:intro}
\end{figure}

\section{Introduction}
\label{sec:intro}
\IEEEPARstart{S}{elf-supervised} learning (SSL) has emerged as a powerful paradigm for learning robust representations from unlabeled data, thereby reducing the reliance on the large-scale, high-quality annotations required by conventional deep learning models \cite{ye2024continual,ye2022desd,ye2025cads,zhou2021models,tang2022self,he2023geometric}. This capability is particularly advantageous in medical image analysis, especially for segmentation tasks, where expert annotation is notoriously labor-intensive and time-consuming \cite{azad2024medical,zhang2021dodnet}.

In medical imaging, SSL methods typically follow a \textbf{pretrain-and-finetune} paradigm to adapt large foundation models to specific clinical applications. The fine-tuning stage tailors pre-trained representations to diverse downstream tasks. Conventionally, this is performed using \textbf{parallel fine-tuning} \cite{ye2024continual,ye2022desd,ye2025cads,zhou2021models,tang2022self,he2023geometric,dong2024efficient,chen2024conv,huang2024fine,tian2024hydralora} (see Fig. \ref{fig:intro}(a)), where a separate model is independently trained for each downstream task starting from the same initial weights. This parallel strategy is suboptimal because it treats each task in isolation, failing to leverage potential synergies and shared knowledge among related clinical tasks. Consequently, it struggles to bridge the representational gap between the general pre-training domain and specific downstream requirements.

To address this limitation, \textbf{multi-task fine-tuning} \cite{agiza2024mtlora,mantri2025} and \textbf{prompt-based learning} \cite{ye2023uniseg,ye2024meduniseg,zhang2021dodnet,liu2023ccq,gao2024training} have been explored to jointly learn from multiple datasets, to exploit shared anatomical and contextual features (see Fig. \ref{fig:intro}(b)). While these approaches can improve performance, they introduce two significant challenges. First, optimization conflicts between competing task objectives can degrade per-task performance on individual tasks. Second, and more critically for clinical deployment, it requires simultaneous access to all task-specific datasets, limiting its flexibility and scalability when new tasks are incrementally introduced.
A more pragmatic approach, which better reflects real-world clinical workflows, is \textbf{sequential fine-tuning}. In this paradigm, a model is progressively adapted to new tasks as they become available, facilitating continuous knowledge transfer and gradual specialization. 

Motivated by this, we propose \textbf{MedSeqFT}, a novel sequential fine-tuning framework for 3D medical image segmentation (Fig. 1c). MedSeqFT reformulates the adaptation of foundation models into a sequential learning pipeline, where a model refined on one task serves as the initialization for the next. This progressive adaptation, however, introduces a critical challenge: balancing task-specific specialization with the retention of previously learned knowledge to avoid \textbf{catastrophic forgetting}. To overcome this challenge, MedSeqFT introduces two key designs: \textbf{Maximum Data Similarity (MDS) selection} and \textbf{Knowledge and Generalization Retention Fine-Tuning (K\&G RFT)}. MDS preserves the model’s generalized knowledge by identifying and buffering downstream samples that are most representative of the original pre-training distribution. K\&G RFT then employs these samples in a novel LoRA-based knowledge distillation (KD) process, which allows the model to learn new task-specific features without catastrophically overwriting its pre-trained weights. 
We conducted extensive experiments on ten diverse 3D segmentation datasets using two state-of-the-art foundation models: VoCo \cite{wu2024large,wu2024voco} and UniMiSS+ \cite{xie2024unimiss+}. Our results demonstrate that MedSeqFT consistently outperforms existing fine-tuning strategies, achieving substantial performance gains (e.g., an average 3.0\% Dice gain). Furthermore, evaluations on two unseen segmentation tasks confirm that the MedSeqFT-refined models exhibit significantly enhanced transferability. This work establishes sequential fine-tuning as a flexible, effective, and knowledge-retentive paradigm for adapting foundation models in medical imaging. The main contributions of this work are:

\begin{itemize}
    \item We propose MedSeqFT, a novel sequential fine-tuning framework that progressively adapts foundation models to a sequence of 3D medical segmentation tasks, offering a flexible and clinically-aligned alternative to conventional fine-tuning paradigms.
    \item We introduce two key components, MDS and K\&G RFT, to explicitly mitigate catastrophic forgetting by preserving core pre-trained knowledge while integrating new task-specific features.
    \item We demonstrate the superiority of MedSeqFT through extensive validation on \textbf{12 diverse CT and MRI datasets}, showing consistent performance gains over existing strategies and significantly enhanced model transferability to unseen tasks.
\end{itemize}

\begin{figure*}[t]
  \centering
  \includegraphics[width=0.95\linewidth]{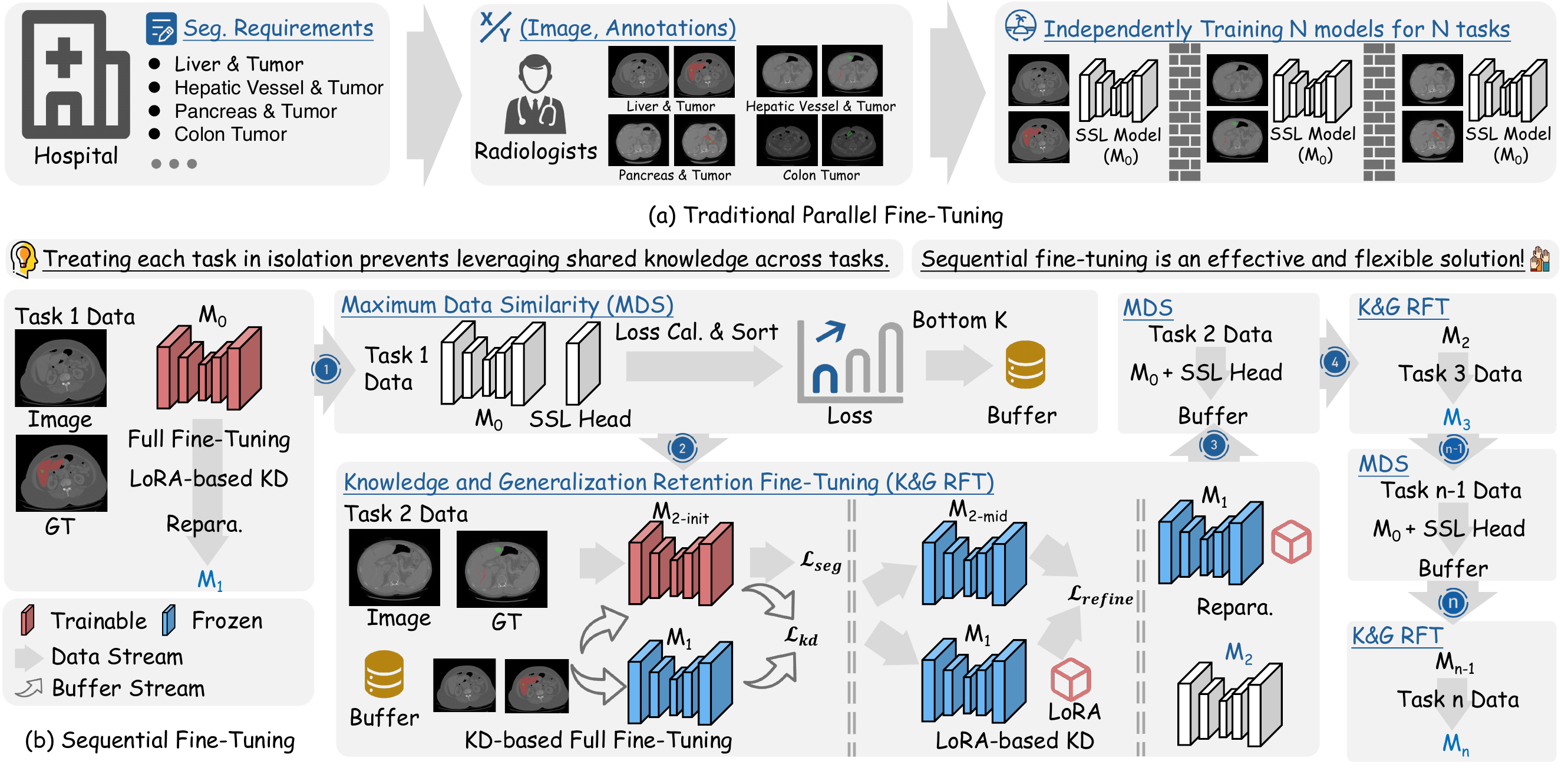}
   \caption{\textbf{Pipeline of the proposed MedSeqFT framework.} MedSeqFT adopts a sequential fine-tuning strategy to progressively adapt a pre-trained model to multiple downstream segmentation tasks while continually refining its representations. To preserve strong generalization across tasks, we introduce two key components: the Maximum Data Similarity (MDS) strategy and the Knowledge and Generalization Retention Fine-Tuning (K\&G RFT) strategy, which jointly mitigate catastrophic forgetting and support robust knowledge transfer.
   }
\label{fig:overview}
\end{figure*}

\section{Related Work}
The ``Pretrain-and-finetune'' paradigm is central to applying foundation models in medical image analysis. However, its effectiveness is often limited by the gap between the generalist pre-training objectives and the specialist demands of downstream tasks. Research to bridge this gap has largely proceeded along two fronts: designing more aligned pre-training tasks and developing more effective fine-tuning strategies.

\subsection{SSL for Better Task Alignment}
One approach to improve downstream performance is to refine the pre-training stage itself \cite{he2022masked}. 
A common strategy is to pre-train on larger, more diverse datasets \cite{zhu2022uni,ye2024continual,wu2024large} to enhance the general quality of learned representations. A more targeted approach involves designing pretext tasks that are better aligned with downstream objectives. For instance, structured knowledge-guided Masked Image Modeling (MIM)\cite{son2024sg} has been shown to improve dense prediction, while MinMax learning \cite{wang2024task} can enhance model robustness for downstream tasks. Other works have tailored pre-training specifically for detection \cite{li2023aligndet}, multi-modal representation \cite{zhang2024unified}, or visual reasoning \cite{wan2022bridging}.
To bridge the gap between upstream and downstream tasks, some methods design task-oriented pretext tasks. For example, structured knowledge-guided MIM enhances dense prediction \cite{son2024sg}, while \textit{MinMax} learning improves robustness for downstream objectives \cite{wang2024task}. Other works tailor pre-training for detection \cite{li2023aligndet}, multi-modal representation \cite{zhang2024unified}, or visual reasoning tasks \cite{wan2022bridging}.
Another line of work infuses pre-training with information about the target task. Methods like Conditional Data Filtering \cite{chakraborty2022efficient} select relevant subsets from the pre-training corpus based on the target domain, while Reference Task Learning \cite{xie2024refs} introduces an auxiliary supervised task to guide representation learning toward the downstream objective. Although these methods can reduce the task gap, they often result in models that are highly specialized, sacrificing the general-purpose utility that makes foundation models attractive.

We contend that refining a general-purpose model for a specific domain, like medical segmentation, is more effectively and flexibly achieved during the fine-tuning process. This motivates our design of MedSeqFT, which focuses on task-specialized model refinement through a sequential, flexible, and knowledge-retentive paradigm.

\subsection{Task Adaptation of Foundation Models}
The fine-tuning strategy is critical for adapting a pre-trained foundation model to target tasks. Existing approaches can be broadly categorized as parallel or multi-task fine-tuning.
\textbf{Parallel fine-tuning} adapts a model to each downstream task independently, with no information shared between tasks. This includes full fine-tuning (FFT) and parameter-efficient fine-tuning (PEFT).
FFT updates all model parameters, including those from the pre-trained backbone and newly added modules (e.g., decoders or segmentation heads) \cite{ye2024continual,ye2022desd,ye2025cads,zhou2021models,tang2022self,haghighi2022dira,he2023geometric}. For example, AC-Norm \cite{zhang2023ac} introduces affine collaborative normalization to dynamically recalibrate non-corresponding channels between pre-training and target tasks, thereby improving channel alignment during fine-tuning. 
PEFT freezes most pre-trained weights and updates only a small subset of parameters. PEFT methods, such as head fine-tuning, Adapters\cite{houlsby2019parameter}, and LoRA \cite{hu2022lora}, have been extended with convolutional designs. For instance, Conv-Adapter \cite{chen2024conv} utilizes depth-wise separable convolutions within a bottleneck structure to preserve spatial locality, Med-Adapter \cite{shenmed}, developed under the Med-Tuning framework, captures both multi-scale spatial and volumetric slice-level correlations via local and frequency-domain features, and Mona \cite{yin20255} further enhances PEFT by incorporating convolutional filters and scaled LayerNorm to adaptively reshape input distributions. While efficient, these parallel methods treat each task in isolation, forgoing the opportunity to share knowledge.
\textbf{Multi-task fine-tuning} addresses this limitation by jointly training on multiple tasks with shared parameters. Recent work has integrated PEFT into this paradigm; for example, MTLoRA \cite{agiza2024mtlora} uses both task-agnostic and task-specific low-rank modules, while DITASK \cite{mantri2025} preserves the singular vectors of weight matrices and adapts only the singular values via neural diffeomorphic transformations, enabling task-specific modulation while maintaining original feature directions. Universal medical segmentation models \cite{ye2023uniseg,ye2024meduniseg,zhang2021dodnet,liu2023ccq,gao2024training} like UniSeg \cite{ye2023uniseg} extend this concept by using a single architecture with prompt-based mechanisms to handle multiple tasks simultaneously. However, these multi-task approaches are inherently inflexible, as they require access to all datasets concurrently and cannot easily scale to accommodate new tasks.


The existing landscape thus presents a trade-off: parallel methods offer flexibility but no knowledge sharing, while multi-task methods share knowledge at the cost of flexibility. Our work, MedSeqFT, proposes a sequential fine-tuning paradigm to resolve this dilemma. It enables progressive knowledge transfer across an expanding set of tasks, combining the adaptability of parallel tuning with the collaborative learning benefits of multi-tasking.

\section{Method}
\subsection{Problem Definition}
Given $n$ datasets $D_1, D_2, \dots, D_n$, each corresponding to a distinct segmentation task, a pre-trained foundation model $M_0$ can be fine-tuned independently on each dataset. The parallel fine-tuning strategy often results in suboptimal performance due to two primary factors: (1) the domain gap between pre-training and fine-tuning tasks, which limits the transferability of pre-trained representations; and (2) the performance degradation when using the strategy of independent fine-tuning on each dataset, due to neglecting the potentially shared information among datasets. To address these limitations, we reframe the task as a sequential learning problem and thus propose MedSeqFT, a framework that progressively refines the foundation model across the sequence of tasks. The goal is to produce a series of models, ${M_1,M_2,\dots,M_n}$, where each model $M_t$ is adapted for task $D_t$ while inheriting and enriching the knowledge from all preceding tasks. This approach enhances task-specific performance and bridges the domain gap by using previously fine-tuned models from preceding tasks as the base instead of generic foundation models, while also enabling knowledge sharing across sequential tasks.



\subsection{Sequential Fine-tuning Framework}
The MedSeqFT pipeline (see Fig. \ref{fig:overview}) executes a sequential refinement process. Based on empirical evidence showing its superior performance, we employ an FFT strategy as the base adaptation mechanism rather than parameter-efficient alternatives. The process unfolds as follows:

\noindent \textbf{1. Initialization (Task 1):} \\
For the first task, $D_1$, the foundation model $M_0$ is fully fine-tuned, followed by LoRA-based KD and Reparameterization (introduced in the K\&G RFT strategy), to produce the first specialized model, $M_1$.

\noindent \textbf{2. Data Selection:} \\
Concurrently, our MDS strategy (Sec. \ref{method:MDS}) is used to select and store a small subset of representative samples from $D_1$ into a buffer, $B_1$.

\noindent \textbf{3. Sequential Refinement (Task $t > 1$):}\\
For each subsequent task $D_t$, the previously refined model, $M_{t-1}$, serves as the new starting point. The model is trained on $D_t$ using our proposed (K\&G RFT) strategy (Sec. \ref{method:KG RFT}), which utilizes the buffered data from previous tasks.

\noindent \textbf{4. Iteration:} \\
After training on $D_t$ is complete, the MDS strategy is again applied to select representative samples from $D_t$, which are added to the buffer. The resulting model, $M_t$, then becomes the pre-trained model for task $D_{t+1}$. This iterative process continues for all $n$ tasks.


\subsection{MDS Strategy} \label{method:MDS}
A key challenge in sequential learning is the potential degradation of the model's general representational capabilities as it adapts to a series of specific tasks. To mitigate this, the MDS strategy aims to preserve a memory of the original pre-training data distribution.
The core principle is that samples from a downstream dataset that yield a very low loss under the original self-supervised pretext task are likely to be structurally similar to the data the model was pre-trained on. 

Specifically, for each dataset $D_t$, we pass all training samples through the initial foundation model $M_0$ and its corresponding SSL head. We compute the SSL loss for each sample; to ensure robust loss values against the stochasticity inherent in SSL tasks (e.g., random masking), this computation is averaged over multiple runs (1000 runs in this study). The $K$ samples with the lowest average SSL loss are identified as being most "similar" to the pre-training distribution and are stored in the buffer for use in the next fine-tuning stage.

 
\subsection{Knowledge and Generalization Retention Fine-Tuning} \label{method:KG RFT}
For all tasks after the first, we employ the K\&G RFT strategy to integrate new task-specific knowledge while explicitly preventing the loss of previously learned representations. This strategy consists of three steps:

\noindent \textbf{1. KD-based FFT:} \\
The model $M_{t-init}$ consists of a vision encoder $E_{t-init}$ initialized from $E_{t-1}$, a decoder $DE_{t-init}$ initialized from $DE_{t-1}$, and a randomly initialized segmentation head. We train $M_{t-init}$ on dataset $D_t$ via FFT, optimizing the segmentation loss $L_{seg}$, which is defined as the sum of Dice loss and cross-entropy loss. Meanwhile, we freeze the previous encoder $E_{t-1}$ and feed buffer samples selected by MDS into both $E_{t-init}$ and $E_{t-1}$. A knowledge distillation (KD) loss, $L_{kd}$ (mean squared error, MSE), is applied to the outputs, encouraging $E_{t-init}$ to retain the representational characteristics of $E_{t-1}$. This preserves knowledge from earlier tasks while maintaining general representational capacity, acting as a soft constraint against catastrophic forgetting. Note that both losses are computed independently to avoid optimization conflicts, with each updating only its corresponding parameters.

\noindent \textbf{2. LoRA-based KD:}\\
KD-based FFT changes $M_{t-init}$ to $M_{t-mid}$ to address task $t$ while inevitably alters the weights of $E_{t-init}$. To provide a hard constraint that isolates and preserves the core pre-trained knowledge, we perform a second KD step. We freeze both $E_{t-1}$ and the newly trained $E_{t-mid}$. Multiple LoRA-based learnable linear layers are then inserted into every linear layer of $E_{t-1}$. We then distill the knowledge from the fine-tuned encoder $E_{t-mid}$ into these LoRA modules using a refinement loss, $L_{refine}$ (MSE), on the new task data. This efficiently captures the `residual' knowledge required for the new task in the low-rank LoRA modules without modifying the original backbone weights. 

\noindent \textbf{3. Reparameterization:}\\
The trained LoRA modules are merged back into the weights of encoder $E_{t-1}$ via reparameterization. A linear layer's weight matrix $W$ is updated as:
\begin{equation}
W' = W + \Delta W = W + BA,
\end{equation}
where $W \in \mathbb{R}^{d \times k}$ is the pre-trained weight, $d$ is the output feature dimension, $k$ denotes the input feature dimension, and $A \in \mathbb{R}^{r \times k}$ and $B \in \mathbb{R}^{d \times r}$ are the low-rank matrices from the LoRA module, with rank $r \ll \min(d,k)$. This step produces the final refined model, $M_t$, which now incorporates knowledge from task $D_{t-mid}$ without increasing its parameter count or architectural complexity, making it ready to serve as the initialization for the next task.
As shown in our visualizations (Sec. \ref{exp:parameter changes}), this hard constraint results in minimal updates to a small subset of the model's parameters. The reparameterization capability of LoRA ensures that the model maintains structural efficiency and avoids redundancy, enabling iterative refinement without increasing model complexity.


\begin{table}[t]
  \centering
  \caption{Number of the training set and test set for each dataset.}
   \resizebox{0.90\columnwidth}{!}{
    \begin{tabular}{cccc}
    \toprule
    \multicolumn{2}{c}{Dataset} & \multicolumn{1}{l}{Train} & \multicolumn{1}{l}{Test} \\
    \midrule
    \multirow{5}[2]{*}{CT Fine-tuning} & Liver & 104   & 27 \\
          & HepaV & 242   & 61 \\
          & Pancreas & 224   & 57 \\
          & Colon & 100   & 26 \\
          & Lung  & 50    & 13 \\
    \midrule
    \multirow{5}[2]{*}{MR Fine-tuning} & PanSegData & 308   & 77 \\
          & ISLES 2022	 & 200   & 50 \\
          & HNTSMRG 2024 & 120   & 30 \\
          & ISPY1 & 128   & 33 \\
          & WMH   & 60    & 110 \\
    \midrule
    \multirow{2}[2]{*}{Refinement Fine-tuning} & COVID-19-20 & 159   & 40 \\
          & Kidney & 168   & 42 \\
    \bottomrule
    \end{tabular}%
    }
  \label{tab: dataset}%
\end{table}%

\begin{table*}[t]
  \centering
  \caption{Performance of TFS, two FFT strategies (the first block), six efficient fine-tuning strategies (the second block), two multi-task fine-tuning strategies (the third block), and our MedSeqFT (the last block) on \textbf{five CT segmentation tasks}. \textbf{VoCo} is the backbone for all strategies. Dice and HD95 are reported as metrics. $\uparrow$: The higher, the better; $\downarrow$: The lower, the better. For LoRA-based strategies, $r$ is rank; for the Adapter, $\gamma$ is a compression factor denoting the down-sampling in the channel dimension. Average scores across five datasets are reported. The best results are highlighted with \textbf{bold}.
  }
    \begin{tabular}{p{0.15\linewidth}p{0.04\linewidth}p{0.04\linewidth}p{0.04\linewidth}p{0.04\linewidth}p{0.04\linewidth}p{0.04\linewidth}p{0.04\linewidth}p{0.04\linewidth}p{0.04\linewidth}p{0.04\linewidth}p{0.04\linewidth}p{0.04\linewidth}}
    \toprule
    \multirow{2}[4]{*}{Method} & \multicolumn{2}{c}{Liver} & \multicolumn{2}{c}{HepaV} & \multicolumn{2}{c}{Pancreas} & \multicolumn{2}{c}{Colon} & \multicolumn{2}{c}{Lung} & \multicolumn{2}{c}{Average} \\
\cmidrule{2-13}          & Dice  & HD95  & Dice  & HD95  & Dice  & HD95  & Dice  & HD95  & Dice  & HD95  & Dice$\uparrow$  & HD95$\downarrow$ \\
    \midrule
    TFS   & 75.6  & 30.6  & 63.5  & 39.3  & 60.2  & 36.4  & 28.8  & 229.5 & 60.0  & 88.6  & 57.6  & 84.9 \\
    FFT & 76.5  & \textbf{19.5}  & 68.7  & 22.5  & 67.2  & 17.0  & 42.8  & 182.5 & 71.5  & 60.6  & 65.3  & 60.4 \\
    ACNorm \cite{zhang2023ac} & 75.4  & 25.4  & 68.6  & 24.7  & 68.3  & 13.6  & 40.9  & 173.0 & 73.3  & 62.5  & 65.3  & 59.9 \\
    \midrule
    Head Fine-tuning & 72.8  & 36.9  & 64.4  & 30.0  & 63.1  & 17.7  & 33.1  & 198.1 & 69.4  & 60.6  & 60.6  & 68.7 \\
    LoRA ($r$=2)  \cite{hu2022lora}   & 73.4  & 33.0  & 65.6  & 30.8  & 65.5  & 13.4  & 38.1  & 198.7 & 68.4  & 52.1  & 62.2  & 65.6 \\
    LoRA ($r$=4)  \cite{hu2022lora}  & 74.0  & 26.2  & 65.0  & 26.5  & 65.1  & 17.1  & 39.2  & 180.7 & 71.8  & 36.6  & 63.0  & 57.4 \\
    LoRA ($r$=8)  \cite{hu2022lora}  & 73.9  & 30.1  & 65.7  & 26.6  & 66.1  & 14.8  & 40.3  & 186.6 & 70.3  & 38.8  & 63.3  & 59.4 \\
    LoRA ($r$=16)  \cite{hu2022lora}  & 74.9  & 30.9  & 66.0  & 25.2  & 65.8  & 15.5  & 38.5  & 203.9 & 75.2  & 36.7  & 64.1  & 62.5 \\
    LoRA ($r$=32)  \cite{hu2022lora}  & 73.4  & 31.9  & 66.8  & 25.7  & 64.0  & 14.9  & 41.3  & 173.4 & 73.9  & 33.8  & 63.9  & 55.9 \\
    Adapter ($\gamma$=2) \cite{houlsby2019parameter} & 73.0  & 34.6  & 65.3  & 29.2  & 65.4  & 17.2  & 38.8  & 193.6 & 67.7  & 88.5  & 62.0  & 72.6 \\
    Adapter ($\gamma$=4) \cite{houlsby2019parameter} & 73.1  & 31.4  & 65.4  & 34.9  & 64.6  & 16.4  & 38.5  & 182.5 & 69.8  & 44.3  & 62.3  & 61.9 \\
    Adapter ($\gamma$=8) \cite{houlsby2019parameter} & 73.5  & 31.5  & 65.0  & 27.5  & 63.5  & 16.4  & 39.1  & 196.6 & 68.1  & 46.7  & 61.9  & 63.7 \\
    Conv-Adapter \cite{chen2024conv} & 75.2  & 29.0  & 67.0  & 22.6  & 67.5  & \textbf{12.1}  & 43.5  & 168.3 & 74.3  & 32.9  & 65.5  & 53.0 \\
    Med-Adapter \cite{shenmed} & 73.3  & 30.3  & 65.9  & 25.2  & 64.1  & 17.5  & 39.2  & 192.9 & 70.2  & 73.0  & 62.5  & 67.8 \\
    Mona \cite{yin20255}  & 74.1  & 32.5  & 64.8  & 40.2  & 65.0  & 16.7  & 39.2  & 205.3 & 69.0  & 41.6  & 62.4  & 67.3 \\
    \midrule
    MTLoRA ($r$=16) \cite{agiza2024mtlora} & 73.9  & 27.5  & 66.2  & 27.1  & 64.5  & 15.6  & 41.1  & 184.6 & 72.3  & 17.8  & 63.6  & 54.5 \\
    MTLoRA ($r$=32) \cite{agiza2024mtlora} & 74.1  & 32.2  & 66.5  & 26.2  & 64.5  & 16.9  & 41.5  & 175.3 & 70.9  & 18.4  & 63.5  & 53.8 \\
    MTLoRA ($r$=64) \cite{agiza2024mtlora} & 74.6  & 28.6  & 66.7  & 28.3  & 63.6  & 14.8  & 43.7  & 162.1 & 71.1  & \textbf{20.2}  & 64.0  & 50.8 \\
    UniSeg \cite{ye2023uniseg} & \textbf{77.7}  & 26.3  & 69.0  & 21.8  & 66.7  & 28.0  & 43.2  & 182.3 & 74.1  & 59.2  & 66.1  & 63.5 \\
    \midrule
    MedSeqFT & 76.5  & \textbf{19.5}  & \textbf{69.8}  & \textbf{18.8}  & \textbf{69.0}  & 12.7  & \textbf{49.8}  & \textbf{160.1} & \textbf{76.4}  & 40.7  & \textbf{68.3}  & \textbf{50.4} \\
    \bottomrule
    \end{tabular}%
  \label{tab:ct_multitask_voco}%
\end{table*}%

\begin{table*}[t]
  \centering
  \caption{Ablation studies of proposed strategies, \textit{i.e.}, maximize data similarity (MDS) and knowledge and generalization retention fine-tuning (K\&G RFT) on five CT segmentation datasets. The results of FFT and sequential fine-tuning (Seq. FT) are reported as baselines. The best result in each column (except for the liver dataset) is highlighted with \textbf{bold}.}
      \resizebox{1.80\columnwidth}{!}{
    \begin{tabular}{ccccc|cccccccccccc}
    \toprule
          & \multicolumn{4}{c|}{Method} & Liver &       & \multicolumn{2}{c}{HepaV} & \multicolumn{2}{c}{Pancreas} & \multicolumn{2}{c}{Colon} & \multicolumn{2}{c}{Lung} & \multicolumn{2}{c}{Average} \\
    \midrule
    FFT  &Seq. FT & MDS   & Random & K\&G RFT & Dice  & HD95  & Dice  & HD95  & Dice  & HD95  & Dice  & HD95  & Dice  & HD95  & Dice  & HD95 \\
    \midrule
    \checkmark &    &       &       &       &     76.5  & 19.5  & 68.7  & 22.5  & 67.2  & 17.0  & 42.8  & 182.5 & 71.5  & 60.6  & 65.3  & 60.4 \\
        &\checkmark &   &   &      &76.5 	&19.5 	&69.1 	&21.7 	&68.6 	&16.8 	&42.0 	&171.0 	&74.1 	&50.9 	&66.1 	&56.0   \\
     &\checkmark    &       & \checkmark &       &    76.5  & 19.5  & 68.9  & 19.7  & 68.1  & 16.4  & 41.2  & 180.6 & 74.5  & 46.0  & 65.8  & 56.4 \\
    & \checkmark    & \checkmark &       &       &    76.5  & 19.5  & \textbf{69.8}  & 19.5  & \textbf{69.1}  & \textbf{12.7}  & 45.5  & 170.9  & 74.4  & 51.3  & 67.1  & 54.8  \\
     &\checkmark    &       &       & \checkmark    & 76.5  & 19.5  & 69.1  & \textbf{18.2}  & 68.9  & 14.2  & 45.1  & 173.9 & 75.8  & 47.6  & 67.1  & 54.7 \\
    & \checkmark    & \checkmark &       & \checkmark   & 76.5  & 19.5  & \textbf{69.8}  & 18.8  & 69.0  & \textbf{12.7}  & \textbf{49.8}  & \textbf{160.1} & \textbf{76.4}  & \textbf{40.7}  & \textbf{68.3}  & \textbf{50.4} \\
    \bottomrule
    \end{tabular}%
    }
  \label{tab:ablation}%
\end{table*}%

\begin{table*}[t]
  \centering
  \caption{Performance of TFS, two FFT strategies (the first block), six efficient fine-tuning strategies (the second block), two multi-task fine-tuning strategies (the third block), and our MedSeqFT (the last block) on \textbf{five CT segmentation tasks}. \textbf{UniMiSS+} is the backbone for all strategies. Dice and HD95 are reported as metrics. $\uparrow$: The higher, the better; $\downarrow$: The lower, the better. For LoRA-based strategies, $r$ is rank; for the Adapter, $\gamma$ is a compression factor denoted the down-sampling in the channel dimension. Average scores across five datasets are reported. The best results are highlighted with \textbf{bold}.}
    \begin{tabular}   {p{0.15\linewidth}p{0.04\linewidth}p{0.04\linewidth}p{0.04\linewidth}p{0.04\linewidth}p{0.04\linewidth}p{0.04\linewidth}p{0.04\linewidth}p{0.04\linewidth}p{0.04\linewidth}p{0.04\linewidth}p{0.04\linewidth}p{0.04\linewidth}}
    \toprule
   \multirow{2}[4]{*}{Method} & \multicolumn{2}{c}{Liver} & \multicolumn{2}{c}{HepaV} & \multicolumn{2}{c}{Pancreas} & \multicolumn{2}{c}{Colon} & \multicolumn{2}{c}{Lung} & \multicolumn{2}{c}{Average} \\
\cmidrule{2-13}          & Dice  & HD95  & Dice  & HD95  & Dice  & HD95  & Dice  & HD95  & Dice  & HD95  & Dice$\uparrow$  & HD95$\downarrow$ \\
    \midrule
    TFS   & 76.4  & 25.2  & 68.4  & 25.4  & 64.0  & 16.9  & 39.7  & 160.1 & 69.4  & 23.8  & 63.6  & 50.3 \\
    FFT & 77.6  & \textbf{16.2}  & 68.6  & 22.6  & 68.1  & 10.2  & 48.0  & 144.2 & 74.7  & 9.4   & 67.4  & 40.5 \\
    ACNorm \cite{zhang2023ac} & 77.4  & 22.2  & 69.1  & 23.6  & 68.6  & 12.3  & 46.1  & 171.6 & 77.0  & 7.7   & 67.6  & 47.5 \\
    \midrule
    Head Fine-tuning & 72.2  & 32.6  & 64.1  & 29.6  & 57.1  & 22.7  & 35.8  & 185.9 & 62.5  & 22.0  & 58.4  & 58.6 \\
    LoRA (r=2)  \cite{hu2022lora}  & 72.1  & 42.5  & 64.9  & 32.1  & 58.3  & 14.9  & 43.1  & 166.0 & 66.6  & 20.3  & 61.0  & 55.2 \\
    LoRA (r=4)  \cite{hu2022lora}  & 72.8  & 30.9  & 66.7  & 25.9  & 57.5  & 13.2  & 43.1  & 174.3 & 62.5  & 21.6  & 60.5  & 53.2 \\
    LoRA (r=8)  \cite{hu2022lora}  & 72.5  & 36.8  & 66.7  & 22.0  & 58.1  & 14.2  & 46.8  & 154.4 & 67.4  & 19.7  & 62.3  & 49.4 \\
    Adapter (r=2) \cite{houlsby2019parameter} & 72.4  & 29.2  & 65.1  & 28.0  & 57.7  & 14.7  & 41.5  & 170.0 & 65.4  & 21.0  & 60.4  & 52.6 \\
    Adapter (r=4) \cite{houlsby2019parameter} & 72.8  & 30.9  & 65.3  & 25.8  & 58.2  & 18.3  & 41.2  & 174.0 & 64.4  & 20.8  & 60.4  & 54.0 \\
    Adapter (r=8) \cite{houlsby2019parameter} & 72.5  & 36.8  & 66.7  & 22.0  & 58.1  & 14.2  & 40.1  & 170.9 & 63.4  & 21.3  & 60.2  & 53.1 \\
    Conv-Adapter \cite{chen2024conv} & 73.0  & 30.7  & 67.5  & 26.8  & 59.3  & 16.3  & 44.3  & 154.9 & 69.2  & 20.4  & 62.6  & 49.8 \\
    Med-Adapter  \cite{shenmed} & 72.5  & 24.4  & 65.0  & 26.9  & 57.5  & 16.6  & 40.7  & 166.6 & 68.0  & 13.4  & 60.7  & 49.6 \\
    Mona \cite{yin20255}  & 71.8  & 32.6  & 65.7  & 27.2  & 59.3  & 17.9  & 44.5  & 163.8 & 68.2  & 19.5  & 61.9  & 52.2 \\
    \midrule
    MTLoRA (r=16) \cite{agiza2024mtlora}  &74.0 	&22.0 	&66.3 	&25.4 	&59.7 	&13.3 	&43.3 	&127.8 	&71.0 	&10.2 	&62.9 	&39.7   \\
    MTLoRA (r=32) \cite{agiza2024mtlora}  & 73.2  & 28.3  & 67.5  & 25.9  & 59.4  & 14.2  & 46.0  & \textbf{105.6} & 67.6  & 18.9  & 62.8  & 38.6 \\
    MTLoRA (r=64) \cite{agiza2024mtlora}  & 73.3  & 30.8  & 67.7  & 22.0  & 58.6  & 19.0  & 43.5  & 129.3 & 70.8  & 18.9  & 62.8  & 44.0 \\
    UniSeg  \cite{ye2023uniseg} & \textbf{79.3}  & 16.5  & \textbf{70.3}  & \textbf{19.2}  & 68.3  & 10.8  & 49.3  & 121.1 & 77.6  & 8.3   & 69.0  & 35.2 \\
    \midrule
    MedSeqFT & 77.6  & \textbf{16.2}  & 69.7  & 20.3  & \textbf{69.0}  & \textbf{10.1}  & \textbf{53.7}  & 116.0 & \textbf{79.7}  & \textbf{5.2}   & \textbf{69.9}  & \textbf{33.6} \\
    \bottomrule
    \end{tabular}%
  \label{tab:ct_multitask_unimiss}%
\end{table*}%

\begin{table*}[t]
  \centering
  \caption{Performance of TFS, two FFT strategies (the first block), six efficient fine-tuning strategies (the second block), two multi-task fine-tuning strategies (the third block), and our MedSeqFT (the last block) on \textbf{five MRI segmentation tasks}. \textbf{VoCo} is the backbone for all strategies. The best results are highlighted with \textbf{bold}.}
    \begin{tabular}{p{0.15\linewidth}p{0.04\linewidth}p{0.04\linewidth}p{0.04\linewidth}p{0.04\linewidth}p{0.04\linewidth}p{0.04\linewidth}p{0.04\linewidth}p{0.04\linewidth}p{0.04\linewidth}p{0.04\linewidth}p{0.04\linewidth}p{0.04\linewidth}}
    \toprule
    \multirow{2}[4]{*}{Method} & \multicolumn{2}{c}{PanSegData} & \multicolumn{2}{c}{ ISLES 2022} & \multicolumn{2}{c}{ HNTSMRG 2024} & \multicolumn{2}{c}{ ISPY1} & \multicolumn{2}{c}{WMH} & \multicolumn{2}{c}{Average} \\
\cmidrule{2-13}          & Dice  & HD95  & Dice  & HD95  & Dice  & HD95  & Dice  & HD95  & Dice  & HD95  & Dice  & HD95 \\
    \midrule
    TFS   & 75.8  & 29.6  & 78.4  & 10.4  & 43.8  & 71.9  & 70.3  & 27.6  & 78.1  & 5.4   & 69.3  & 29.0 \\
    FFT & \textbf{81.8} & \textbf{11.6} & 78.9  & 10.4  & 55.8  & 49.5  & 71.6  & 20.2  & 77.9  & 5.2   & 73.2  & 19.4 \\
    ACNorm \cite{zhang2023ac} & 81.6  & 9.2   & 79.1  & 12.1  & \textbf{56.5} & \textbf{45.8} & 72.6  & 22.0  & 77.8  & 4.8   & 73.5  & 18.8 \\
    \midrule
    Head Fine-tuning & 78.7  & 20.6  & 59.9  & 25.9  & 52.9  & 47.2  & 55.8  & 43.5  & 65.8  & 9.5   & 62.6  & 29.4 \\
    LoRA (r=2) \cite{hu2022lora} & 79.7  & 17.3  & 61.6  & 20.6  & 53.4  & 53.3  & 57.9  & 28.2  & 66.2  & 9.6   & 63.7  & 25.8 \\
    LoRA (r=4) \cite{hu2022lora}  & 79.9  & 13.6  & 62.6  & 21.4  & 54.1  & 46.4  & 60.0  & 22.5  & 65.9  & 11.2  & 64.5  & 23.0 \\
    LoRA (r=8) \cite{hu2022lora}  & 79.9  & 13.4  & 62.6  & 20.5  & 55.1  & 47.2  & 59.5  & 30.3  & 66.4  & 8.4   & 64.7  & 24.0 \\
    Adapter (r=2) \cite{houlsby2019parameter} & 79.9  & 14.7  & 61.7  & 22.6  & 54.4  & 48.9  & 56.9  & 32.2  & 65.6  & 10.6  & 63.7  & 25.8 \\
    Adapter (r=4) \cite{houlsby2019parameter} & 79.1  & 16.2  & 60.5  & 22.9  & 54.4  & 49.3  & 55.5  & 40.7  & 65.8  & 10.6  & 63.1  & 27.9 \\
    Adapter (r=8) \cite{houlsby2019parameter} & 79.1  & 17.2  & 61.5  & 21.7  & 54.3  & 48.2  & 57.9  & 32.4  & 65.7  & 10.4  & 63.7  & 26.0 \\
    Conv-Adapter \cite{chen2024conv} &79.9	&15.2	&76.9	&10.1	&54.0	&42.8	&66.6	&23.1	&76.4	&7.6	&70.8	&19.8  \\
    Med-Adapter \cite{shenmed} & 79.6  & 14.8  & 60.7  & 22.4  & 53.3  & 49.9  & 57.8  & 31.6  & 64.8  & 11.4  & 63.2  & 26.0 \\
    Mona \cite{yin20255}  & 79.7  & 14.4  & 61.2  & 22.5  & 53.6  & 48.5  & 58.0  & 23.6  & 65.5  & 11.6  & 63.6  & 24.1 \\
    \midrule
    MTLoRA (r=16) \cite{agiza2024mtlora} & 77.9  & 19.5  & 63.3  & 19.9  & 51.1  & 72.1  & 62.6  & 19.3  & 75.1  & 6.8   & 66.0  & 27.5 \\
    MTLoRA (r=32) \cite{agiza2024mtlora} & 78.0  & 17.8  & 60.5  & 24.0  & 51.2  & 73.1  & 62.3  & 26.2  & 75.5  & 6.9   & 65.5  & 29.6 \\
    MTLoRA (r=64) \cite{agiza2024mtlora} & 78.4  & 14.9  & 61.4  & 19.2  & 51.3  & 76.3  & 62.4  & 22.5  & 75.5  & 6.1   & 65.8  & 27.8 \\
    UniSeg \cite{ye2023uniseg} & 79.5  & \textbf{11.6} & 79.2  & 9.3   & 56.0  & 55.2  & 74.2  & \textbf{14.0} & 78.2  & 4.5   & 73.4  & 18.9 \\
    \midrule
    MedSeqFT & \textbf{81.8} & \textbf{11.6} & \textbf{79.6} & \textbf{8.7} & \textbf{56.5} & 50.3  & \textbf{76.6} & 14.9  & \textbf{78.6} & \textbf{4.3} & \textbf{74.6} & \textbf{18.0} \\
    \bottomrule
    \end{tabular}%
  \label{tab:mri_multitask_voco}%
\end{table*}%

\begin{table}[t]
  \centering
  \caption{Performance of TFS, FFT, and FFT using pre-trained models from MedSeqFT on two CT datasets. The best results are highlighted with \textbf{bold}.}
    \begin{tabular}{ccccccc}
    \toprule
    \multicolumn{1}{c}{\multirow{3}[6]{*}{Method}} & \multicolumn{2}{c}{\multirow{2}[4]{*}{COVID-19-20}} & \multicolumn{4}{c}{Kidney} \\
\cmidrule{4-7}    \multicolumn{1}{c}{} & \multicolumn{2}{c}{} & \multicolumn{2}{c}{Kidney} & \multicolumn{2}{c}{Tumor} \\
\cmidrule{2-7}    \multicolumn{1}{c}{} & Dice  & HD95  & Dice  & HD95  & Dice  & HD95 \\
    \midrule
    TFS   & 53.8  & 91.1  & 90.7  & 29.9  & 29.4  & 180.2 \\
    FFT   & 65.5  & 54.2  & \textbf{96.4}  & \textbf{4.0}   & 68.5  & 91.6 \\
    FFT w/ MedSeqFT & \textbf{66.8}  & \textbf{45.9}  & 96.3  & 4.5   & \textbf{72.0}  & \textbf{81.9} \\
    \bottomrule
    \end{tabular}%
  \label{tab:refined model}%
\end{table}%

\section{Experiment}
\subsection{Datasets}
We employed three groups of 3D medical image segmentation datasets to comprehensively evaluate the proposed MedSeqFT framework. Two groups were used to assess multi-task fine-tuning performance, and the third was used to examine the transferability of the refined pre-trained model.

\textbf{Multi-task Fine-tuning Sets:} 
We curated ten publicly available datasets, divided by imaging modality into CT and MRI groups.
The \textbf{CT multi-task fine-tuning set} consists of five datasets: the Liver dataset, derived from LiTS \cite{bilic2023liver}, which provides liver and tumor annotations; and four datasets from the Medical Segmentation Decathlon (MSD) Challenge \cite{antonelli2022medical}, including the HepaV dataset (hepatic vessels and tumors), the Pancreas dataset (pancreas and tumors), the Colon dataset (colon tumors), and the Lung dataset (lung tumors).
The \textbf{MRI multi-task fine-tuning set} includes PanSegData, ISLES 2022, HNTSMRG 2024, ISPY1, and WMH.
The PanSegData dataset \cite{zhang2024large} provides pancreas annotations on both T1-weighted (T1W) and T2-weighted (T2W) MRI scans, comprising 767 scans (385 T1W and 382 T2W) from five centers. We adopted the 385 T1W scans for evaluation. 
The ISLES 2022 dataset \cite{hernandez2022isles} provides ischemic stroke lesion annotations on diffusion-weighted imaging (DWI), apparent diffusion coefficient (ADC), and fluid-attenuated inversion recovery (FLAIR) scans, with 400 cases collected from three centers at 1.5T and 3T. Of these, 250 cases are publicly available for training/validation, and we used them for evaluation. 
The HNTSMRG 2024 dataset \cite{wahid_2024_11199559} contains annotations of head and neck tumors, including primary gross tumor volumes (GTVp) and metastatic lymph nodes (GTVn), on T2W scans. We used the publicly available Task 1 training set. 
The ISPY1 dataset \cite{chitalia2022expert} provides structural breast tumor volumes on three MRI sequences: pre-contrast, first post-contrast, and second post-contrast scans. 
Finally, the WMH dataset \cite{kuijf2019standardized} provides annotations of white matter hyperintensities of presumed vascular origin on brain MRI scans, including both T1W and FLAIR images.

\textbf{Refinement and Transferability Set:} To assess transferability, we employed two CT datasets not included in the multi-task sets. The COVID-19-20 dataset \cite{roth2022rapid} provides annotations of COVID-19 lung lesions, while the Kidney dataset, derived from KiTS \cite{heller2021state}, includes CT scans of kidney cancer patients undergoing nephrectomy, annotated with kidneys and kidney tumors.

Details of the training and testing splits are summarized in Table~\ref{tab: dataset}. For all datasets, an 80:20 ratio was applied for partitioning, except for WMH (we follow the official split).

\subsection{Implementations}
All experiments were implemented using two learning frameworks built upon the nnU-Net architecture \cite{isensee2021nnu}: (1) supervised fine-tuning and (2) LoRA-based KD.

For \textbf{supervised fine-tuning}, we adopted two state-of-the-art CT-pretrained models as backbones: VoCo (Swin-Base) \cite{wu2024large,wu2024voco} and UniMiSS+ \cite{xie2024unimiss+}. We followed the official configurations for both models, training for 25,000 iterations with the AdamW optimizer. For VoCo, the learning rate was set to $3e^{-4}$, the batch size was 2, and the input patch size was $32\times160\times160$. For UniMiSS+, the learning rate was $1e^{-4}$, the batch size was 2, and the input patch size was $32\times160\times160$.

For \textbf{LoRA-based KD}, we set the LoRA rank to 2 to constrain weight modifications and preserve pre-trained knowledge. The learning rate was set to $3e^{-5}$ for VoCo and $1e^{-4}$ for UniMiSS+. For VoCo, the number of distillation iterations was set to 5,000 for the CT multi-task dataset and 20,000 for the MRI multi-task dataset to ensure training convergence. For UniMiSS+, the distillation process was conducted for 12,500 iterations. All other hyperparameters followed the standard settings used in the corresponding frameworks.

\subsection{Evaluation Metrics}
Model performance was evaluated using two standard metrics: the Dice Similarity Coefficient (Dice, \%) and the 95th Percentile of the Hausdorff Distance (HD95). For datasets with multiple classes, the reported score is the mean value of the metric across all categories.

\begin{figure}[t]
  \centering
  \includegraphics[width=0.95\linewidth]{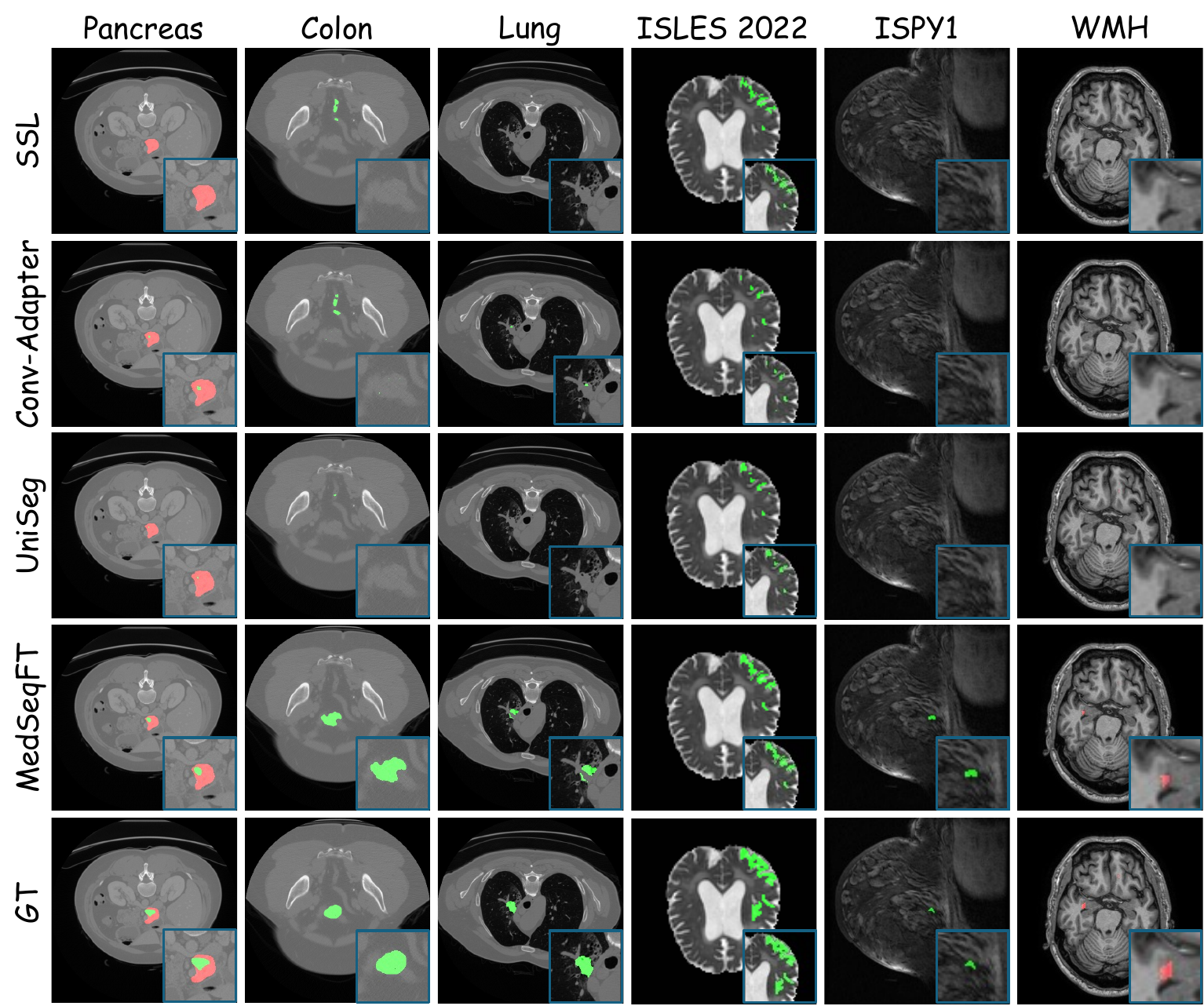}
   \caption{Visualization of segmentation results obtained by SSL, Conv-Adapter, UniSeg, MedSeqFT, and ground truths (GTs) on six datasets. The lesion and non-lesion regions are colored green and red, respectively. The enlarged region is displayed in the right corner of each image.
   }
\label{fig:visual_voco}
\end{figure}

\subsection{Comparing to Fine-tuning strategies} \label{exp:voco fine-tuning}
We compared our proposed MedSeqFT sequential fine-tuning strategy against a wide range of existing fine-tuning approaches on five CT segmentation datasets. For a fair comparison, all methods used the same pre-trained VoCo backbone and were trained for an equal number of iterations. The competing strategies include two FFT approaches (vanilla FFT and ACNorm \cite{zhang2023ac}), six PEFT approaches (head fine-tuning, LoRA \cite{hu2022lora}, Adapter \cite{houlsby2019parameter}, Conv-Adapter \cite{chen2024conv}, Med-Adapter \cite{shenmed}, and Mona \cite{yin20255}), and two multi-task fine-tuning approaches (MTLoRA \cite{agiza2024mtlora} and UniSeg \cite{ye2023uniseg}). 
The results on the CT multi-task fine-tuning benchmark, which integrates five individual segmentation tasks, are summarized in Table~\ref{tab:ct_multitask_voco} and give rise to several key observations. 
First, among FFT strategies, SSL initialization consistently outperforms training from scratch (TFS), with average improvements of 7.7\% in Dice and 24.5 mm in HD95, highlighting the importance of pre-training.
Second, although PEFT methods update substantially fewer parameters, they generally exhibit inferior performance compared to FFT. Among them, Conv-Adapter delivers the strongest results, closely approaching the performance of FFT. However, this parity does not persist when applied to the UniMiSS+ model (see Table~\ref{exp:unimiss+}), indicating potential model-dependent limitations. 
Third, among multi-task strategies, UniSeg achieves the highest Dice scores but suffers from degraded HD95 performance, indicating possible adverse effects of joint training. We hypothesize that its task-shared decoder and segmentation head may induce optimization conflicts across tasks — an issue alleviated by strategies such as MTLoRA, which employ task-specific decoders.
Fourth, our MedSeqFT strategy achieves the best Dice score across all five datasets and the best HD95 score on three of them. On average, MedSeqFT improves Dice by 3.0\% and HD95 by 10 mm compared to conventional FFT, ranking highest on both metrics. These results collectively demonstrate the superior effectiveness and robustness of MedSeqFT in adapting SSL models to diverse CT segmentation tasks.


\subsection{Ablation Studies}
We conducted ablation studies on the CT multi-task dataset to evaluate the effectiveness of the proposed MDS and K\&G RFT components. The results are compared against two baselines: the standard FFT and vanilla sequential fine-tuning (Seq. FT), where the model trained on the previous dataset directly initializes training for the next (except that the segmentation head is randomly initialized). As all variants apply FFT to the first task (Liver dataset), their performance on this dataset is identical.
The results, summarized in Table \ref{tab:ablation}, support the following conclusions:
First, compared to FFT, the vanilla Seq. FT baseline achieves better performance on most metrics, suggesting an inherent advantage to sequential training in multi-task scenarios.
Second, integrating MDS with Seq. FT consistently improves performance across the remaining four datasets. In a direct comparison with a strategy that randomly selects buffered data (of the same size), MDS yields superior results in both average Dice and HD95, demonstrating the benefit of strategically preserving the pre-trained model’s representation capacity.
Third, Seq. FT with K\&G RFT also shows clear benefits over the vanilla Seq. FT baseline, achieving performance comparable to that of Seq. FT with MDS. This indicates the effectiveness of our KD and generalization retention mechanisms in mitigating catastrophic forgetting.
Fourth, the complete MedSeqFT framework, which combines both MDS and K\&G RFT, yields the best generalization performance across all five datasets (a 3.0\% improvement in Dice and a 10mm reduction in HD95 compared to FFT), highlighting the complementary strengths of the two strategies.

\subsection{Adopting MedSeqFT to UniMiSS+} \label{exp:unimiss+}
To assess the robustness and generalizability of our framework, we replaced the VoCo backbone with UniMiSS+ and repeated the experiments on the CT multi-task dataset. The results are presented in Table \ref{tab:ct_multitask_unimiss}.

The findings are consistent with the observations made using VoCo. The pre-trained UniMiSS+ model provides a strong baseline, validating the benefits of SSL pre-training. PEFT strategies remain suboptimal, exhibiting notable performance degradation compared to FFT. Notably, Conv-Adapter, which performed well with VoCo, fails to match FFT performance with UniMiSS+, showing a 4.8\% drop in Dice. In contrast, MedSeqFT again delivers the best average performance across all metrics, achieving consistent improvements of 2.5\% in Dice and 6.9 mm in HD95 over FFT. These results demonstrate that MedSeqFT is not limited to a specific backbone and generalizes well to other SSL pre-trained models.

\subsection{Adopting MedSeqFT to MRI Multi-task Dataset}
In this experiment, we investigated the performance of MedSeqFT when transferring across modalities. Specifically, we used the VoCo model, which was pre-trained exclusively on CT data, as the initialization for fine-tuning on the five MRI segmentation datasets.
The results are reported in Table \ref{tab:mri_multitask_voco}. Consistent with earlier findings, MedSeqFT outperforms FFT across most metrics, achieving average improvements of 1.4\% in Dice and 1.4 mm in HD95. We also observed that the CT-pretrained model did not always generalize well to MRI tasks under standard fine-tuning. For instance, on the WMH dataset, using the pre-trained weights with FFT did not lead to performance gains over TFS. In contrast, MedSeqFT consistently improved performance across all MRI datasets, including WMH, where it achieved improvements of 0.5\% in Dice and 1.1 mm in HD95 over TFS. These results highlight the robustness and adaptability of MedSeqFT, even when transferring across modalities where significant domain shifts can hinder traditional fine-tuning approaches.

\subsection{Visualization of Segmentation Results}
We qualitatively compared the segmentation results from MedSeqFT with those from SSL (FFT), Conv-Adapter, and UniSeg on six datasets, as illustrated in Fig. \ref{fig:visual_voco}. The visualizations show that MedSeqFT consistently produces segmentations that most closely align with the ground truths (GTs), effectively mitigating issues such as under-segmentation. For instance, in the case of pancreatic tumor segmentation (first column of Fig. \ref{fig:visual_voco}), the SSL, Conv-Adapter, and UniSeg either miss substantial tumor regions or fail to detect them entirely. In contrast, MedSeqFT identifies considerably larger tumor regions, which are more consistent with the ground truth.


\begin{figure*}[t]
  \centering
  \includegraphics[width=0.90\linewidth]{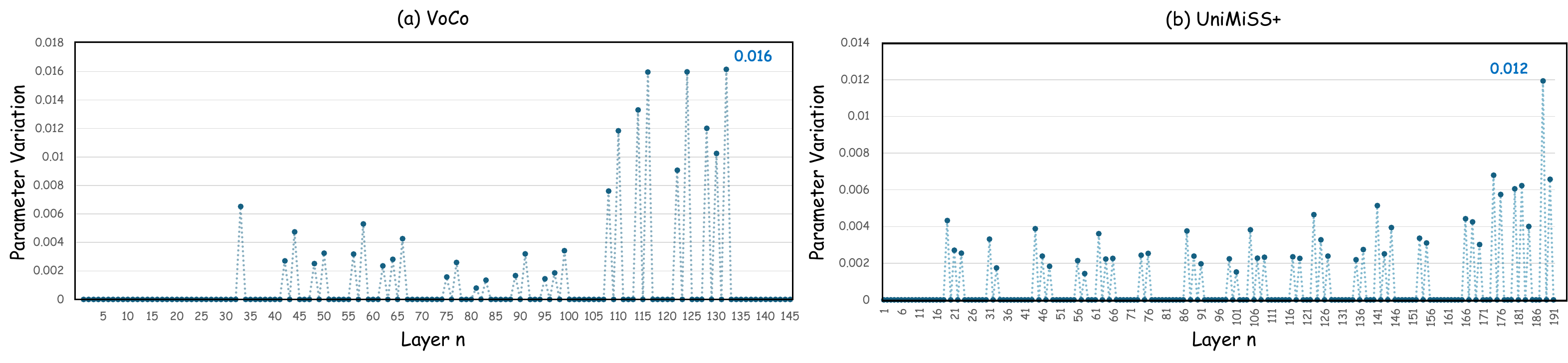}
   \caption{Visualization of parameter variation for (a) VoCo and (b) UniMiSS+ after refinement on the CT multi-task fine-tuning dataset using MedSeqFT. Each point corresponds to a specific encoder layer ($n$ denotes the $n$-th encoder layer). Parameter variation is computed as the average absolute change in parameter values before and after refinement.
   }
\label{fig:param_variation}
\end{figure*}

\begin{figure}[t]
  \centering
  \includegraphics[width=0.85\linewidth]{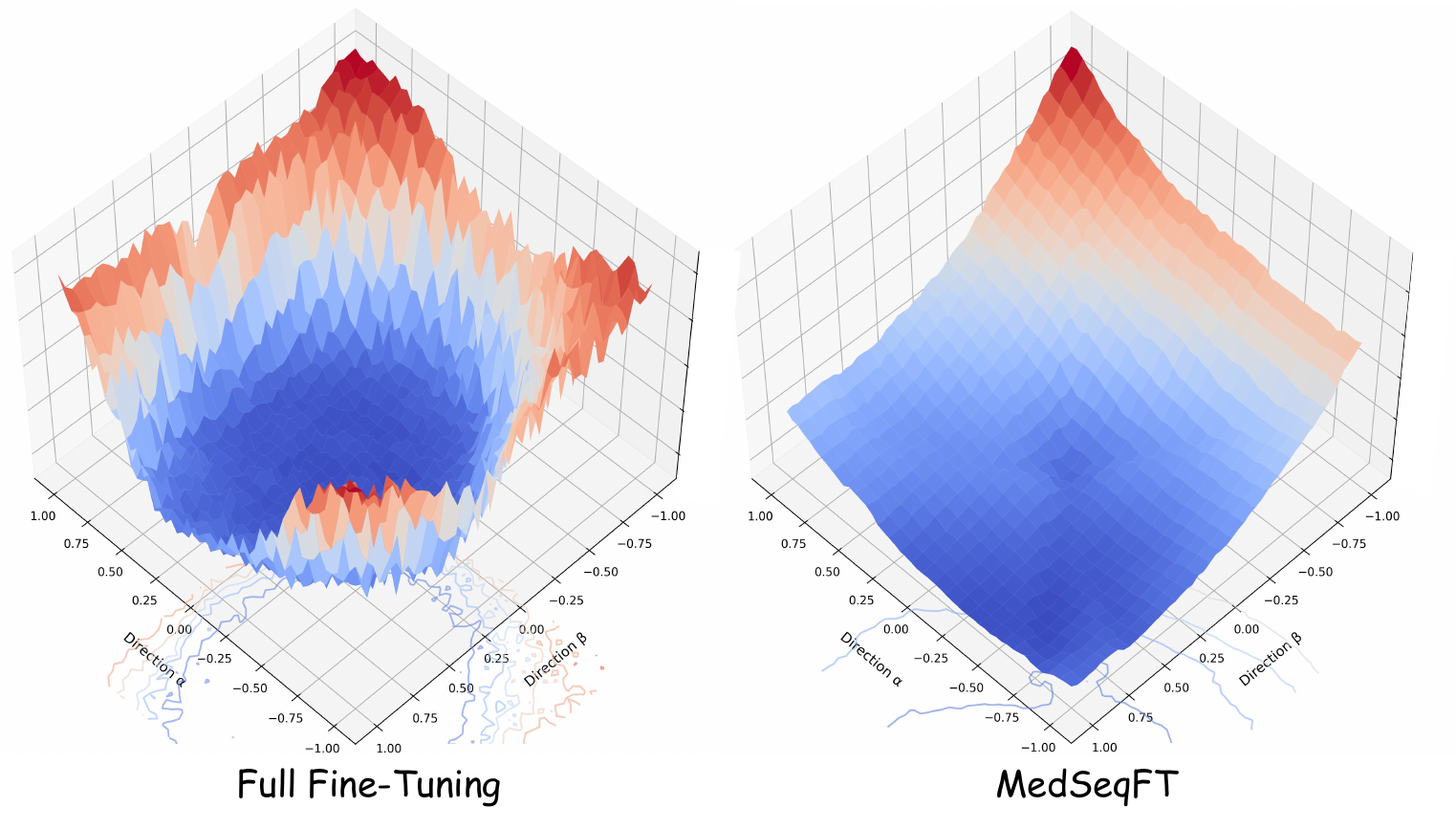}
   \caption{Visualization of the loss landscapes \cite{li2018visualizing} for FFT and MedSeqFT, both trained on the Lung dataset. Colors represent loss values, increasing progressively from blue (low loss) to red (high loss).
   }
\label{fig:loss_landscape}
\end{figure}

\section{Discussion}
\subsection{Effectiveness of Refined Pre-trained Model}
To assess the transferability of the refined pre-trained model, we evaluated performance on two additional CT segmentation datasets: Kidney and COVID-19-20. We compared the original SSL pre-trained model with its refined counterpart produced by MedSeqFT, with both models fine-tuned using the FFT strategy. The results, summarized in Table \ref{tab:refined model}, reveal several noteworthy findings.
The original pre-trained model (FFT) already delivers substantial improvements over TFS across all metrics, demonstrating the broad utility of SSL pre-training. Building upon this, the refined model (FFT w/ MedSeqFT), which was further adapted via MedSeqFT on the CT multi-task dataset, achieves additional performance gains, particularly in tumor segmentation tasks. Specifically, it yields average improvements of 3.5\% in Dice and 9.7 mm in HD95 compared to the original FFT model.
These findings indicate that MedSeqFT effectively tailors the pre-trained model for CT segmentation, thereby enhancing its suitability for downstream deployment. The gains are especially pronounced in tumor segmentation, reflecting the task emphasis incorporated during sequential refinement.
To further characterize the learning dynamics, we visualized the loss landscapes \cite{li2018visualizing} of models trained with FFT and with MedSeqFT on the Lung dataset under a multi-task learning setting (see Fig.~\ref{fig:loss_landscape}). For FFT, the central region of the surface exhibits a low-loss concavity, but the surrounding areas fluctuate sharply, with pronounced oscillations and peaks. In contrast, the surface produced by MedSeqFT is substantially smoother, with more uniform gradient variations and a lack of high-frequency perturbations. These characteristics suggest that models optimized with MedSeqFT achieve greater stability and robustness—properties that are particularly advantageous for transfer learning.

\subsection{Changes of Refined Pre-trained Model} \label{exp:parameter changes}
To further analyze the effect of MedSeqFT, we visualized the parameter variation of the pre-trained encoder before and after refinement on the CT multi-task dataset, as shown in Fig. \ref{fig:param_variation}. The variation was computed by taking the absolute difference between corresponding parameter values and then averaging over all elements in a given layer. We observed that in the VoCo model, the linear layers in shallow encoder blocks showed no variation. This is likely due to the absence of gradient updates in their corresponding LoRA layers during the K\&G RFT stage. In contrast, UniMiSS+ did not exhibit this behavior, indicating a difference in model architecture. From these visualizations, several key findings emerge:
(1) Most parameters remain unchanged after refinement, which helps to preserve the generalization capability of the pre-trained model.
(2) Deeper layers exhibit more variation than shallower ones. This supports the established understanding that deep layers tend to learn more task-specific features, while shallow layers capture more general-purpose representations.
(3) The overall magnitude of parameter changes is minimal, with maximum average variations of only 0.016 for VoCo and 0.012 for UniMiSS+, demonstrating the subtle nature of the refinement process.
In summary, MedSeqFT introduces minimal yet effective parameter updates, which refine the pre-trained model while preserving its generalization ability.

\begin{table}[t]
  \centering
  \caption{Time cost of FFT and MedSeqFT on the CT multi-task dataset. VoCo is the backbone. H: hours.}
\resizebox{1.0\columnwidth}{!}{
    \begin{tabular}{c|cccc}
    \toprule
    Method & \multicolumn{4}{c}{Training Time (5 tasks)} \\
    \midrule
    FFT   & \multicolumn{4}{c}{$\sim$21.1 h} \\
    \midrule
    \multirow{2}[4]{*}{MedSeqFT} & FFT (The first stage) & KD-based FFT & LoRA-based KD & Sum \\
\cmidrule{2-5}          & $\sim$4.2h & $\sim$19.9h & $\sim$2.9h & $\sim$27.0h (+5.9h) \\
    \bottomrule
    \end{tabular}%
    }
  \label{tab:time_need}%
\end{table}%

\subsection{Resource Requirements for MedSeqFT}
Compared to the parallel FFT strategy, our proposed MedSeqFT introduces two additional stages: KD-based FFT and LoRA-based KD. We analyzed the time and GPU memory costs of both FFT and MedSeqFT on the CT multi-task dataset using the VoCo backbone. As shown in Table~\ref{tab:time_need}, FFT requires approximately 21.1 hours to complete training across all five tasks. MedSeqFT requires 27.0 hours, resulting in an additional 5.9 hours of training time. This increase is attributed to the two new stages: (1) the KD-based FFT stage, applied to the last four tasks, adds approximately 3.0 hours, and (2) the LoRA-based KD stage contributes an additional 2.9 hours. 
In terms of GPU memory usage, the KD-based FFT stage incurs minimal overhead, as it only adds a frozen encoder and a KD loss. The LoRA-based KD stage imposes even lower memory requirements since it only updates the lightweight LoRA layers. 
In summary, MedSeqFT introduces a modest increase in training time and a negligible additional memory cost compared to FFT, while achieving significant performance improvements, which makes MedSeqFT a promising and practical fine-tuning strategy for clinical deployment.

\section{Conclusion}
We presented MedSeqFT, a sequential fine-tuning framework for adapting self-supervised foundation models to 3D medical image segmentation. Across ten CT and MRI tasks, MedSeqFT achieved consistent improvements over existing fine-tuning strategies, with an average 3.0\% Dice gain and 10 mm HD95 reduction compared to conventional full fine-tuning. On two unseen datasets, it further enhanced transferability, particularly for tumor segmentation. Analyses of loss landscapes and parameter variations confirmed that MedSeqFT refines models with minimal yet effective updates, resulting in greater stability and robustness. Despite a modest increase in training time, MedSeqFT offers a flexible and clinically aligned paradigm for progressive model adaptation. Future work will explore selective refinement of pre-trained components and extend the framework to multi-modal imaging tasks.

\bibliographystyle{IEEEtranS}
\bibliography{main}

\end{document}